%% file: main.tex
\documentclass[letterpaper, 10 pt, conference]{ieeeconf}

\IEEEoverridecommandlockouts

\usepackage{cite}
\usepackage{amsmath,amssymb,amsfonts}
\usepackage{algorithmic}
\usepackage{graphicx}
\usepackage{textcomp}
\usepackage{xcolor}
\usepackage{dialogue}
\def\BibTeX{{\rm B\kern-.05em{\sc i\kern-.025em b}\kern-.08em
    T\kern-.1667em\lower.7ex\hbox{E}\kern-.125emX}}

\makeatletter
\def\namedlabel#1#2{\begingroup
   \def\@currentlabel{#2}%
   \label{#1}\endgroup
}
\makeatother
     
\makeatletter
\newcommand{\linebreakand}{%
  \end{@IEEEauthorhalign}
  \hfill\mbox{}\par
  \mbox{}\hfill\begin{@IEEEauthorhalign}
}
\makeatother

\begin{document}

\title{\LARGE \bf Into the Wild: When Robots Are Not Welcome}

\author{Shaul Ashkenazi$^{1}$, Gabriel Skantze$^{2}$, Jane Stuart-Smith$^{3}$, and Mary Ellen Foster$^{1}$
\thanks{$^{1}$ School of Computing Science, University of Glasgow, Glasgow, UK
        {\tt\small Shaul.Ashkenazi@glasgow.ac.uk}}%
\thanks{$^{2}$ Division of Speech, Music and Hearing, KTH Royal Institute of Technology, Stockholm, Sweden}%
\thanks{$^{3}$ School of Critical Studies, University of Glasgow, Glasgow, UK}%
}

\maketitle
\thispagestyle{empty}
\pagestyle{empty}

\begin{abstract}
    Social robots are increasingly being deployed in public spaces, where they face not only technological difficulties and unexpected user utterances, but also objections from stakeholders who may not be comfortable with introducing a robot into those spaces. We describe our difficulties with deploying a social robot in two different public settings: 1) Student services center; 2) Refugees and asylum seekers drop-in service. Although this is a failure report, in each use case we eventually managed to earn the trust of the staff and form a relationship with them, allowing us to deploy our robot and conduct our studies.
\end{abstract}

\input{sections/introduction}
\input{sections/related-work}
\input{sections/public-spaces}
\input{sections/conclusions}

\section*{ACKNOWLEDGMENT}

This work was supported by the \textit{UKRI Centre for Doctoral Training in Socially Intelligent Artificial Agents}, Grant Number EP/S02266X/1.

\bibliographystyle{IEEEtran}
\bibliography{references}

\end{document}

%% file: sections/introduction.tex
\section{Introduction}

We have developed a multilingual robot system (Figure ~\ref{fig:robot}) described in \cite{10.1145/3610978.3640633} for two different use cases: 1) Supporting newly arrived international students in a UK university, answering frequently asked questions; 2) Supporting refugees and asylum seekers with navigating bureaucratic processes.

Like most current public-space robot deployments, our field studies involved
adding a robot to an existing workplace, with stakeholders including
management, visitors, as well as front-line workers who should all be
consulted to develop the details of the system to be deployed.
For this sort of public robot deployment to be successful, the support of the other stakeholders is equally important as the responses of the building visitors. In this report, we discuss how we engaged with stakeholders in two very different public spaces, and reflected on our repeated failure (but eventual success) in gaining approval to deploy our robot in the wild.

%% file: sections/related-work.tex
\section{Related Work}

Social robots are being increasingly deployed in public spaces, where the technical challenges are significant, but the application areas and evaluations are more authentic and realistic \cite{mubin2018,schneider2022}. For example, the Furhat robot can be found acting as a receptionist \cite{moujahid2022}, a museum guide \cite{muller2023}, and a barista \cite{lim2023}, while Pepper has been deployed in a wide range of contexts including shopping malls \cite{foster-mummer}, restaurants \cite{stock2018can}, libraries \cite{10.1145/3334480.3382979}, university buildings \cite{blair2023development}, and train stations \cite{10.1145/3371382.3378294}. User responses to these robots have generally been positive, although technical limitations on audiovisual sensing in public spaces sometimes have an impact, and many studies suffer from the novelty effect \cite{reimann2023social}.

\begin{figure}[t!]
    \centerline{\includegraphics[width=\columnwidth]{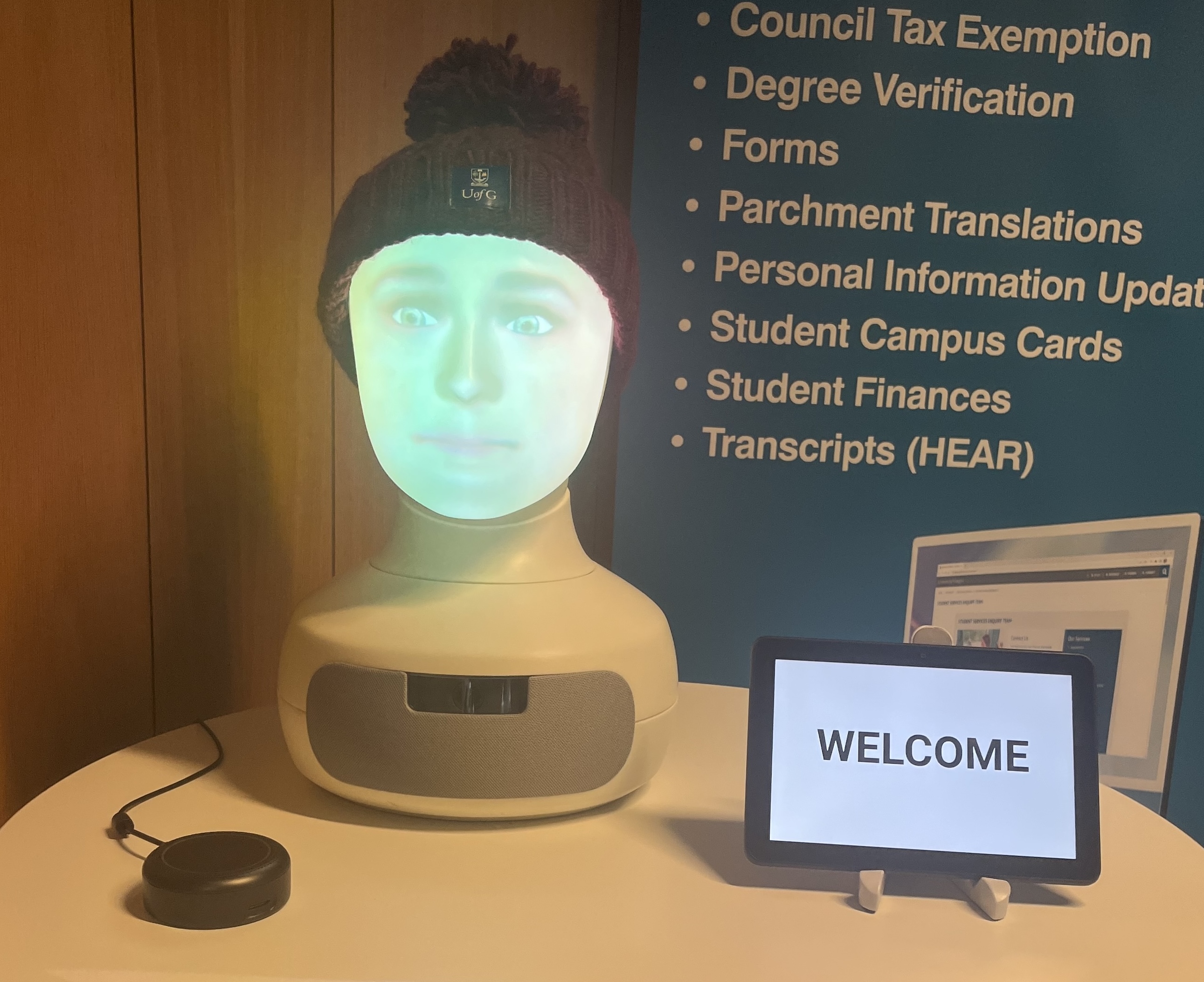}}
    \caption{The robot system}
    \label{fig:robot}
\end{figure}

Deploying a robot in a public space normally involves adding a robot to an
existing workplace (as in all of the deployments listed above), with
stakeholders including management, visitors, as well as front-line workers, who
should all be consulted to develop the details of the system to be deployed
\cite{Niemelä2019}. The managers are clearly influential in this process: no
deployment can ethically be carried out in such a space without the permission
of the management, who also exercise significant control over the time and place
of the deployment, as well as the topics that the robot might discuss and the
way in which it engages with visitors (e.g., \cite{9900717}). On the other hand, the success of most
interactive robot systems is also normally assessed through studying the
behavior and subjective responses of users (i.e., building visitors)
\cite{Apraiz_Lasa_Mazmela_2023}. However, for this sort of public robot
deployment to be successful, the input and feedback from other stakeholders is
equally important---in fact, without feedback from the front-line workers,
no deployment will likely be successful. Some workers may
initially fear being replaced by a robot, but in practice the most productive model
is one where the robot works alongside the humans collaboratively \cite{zhang2024humanlike}, carrying out some tasks and freeing the workers to complete others.

%% file: sections/public-spaces.tex
\section{Public Spaces}

\subsection{Student Services Center}

Our relationship with the student support center staff and management has had its ups and downs, and it took several consultations before the in-the-wild deployment was eventually approved.

\subsubsection{First Iteration}
It took some time and numerous emails to get an initial meeting with one of the
managers, just to discuss the possibility of having a robot deployed in the
student support center. After that, as part of finalizing the behavior of the robot, we sat next to the support officers on several occasions to understand
the common topics students ask about, and what was expected of a
support officer. 

\subsubsection{Second Iteration}
Some time into the project, we were informed that a new center manager 
had not been aware of the project and wanted to understand it better. Following
a meeting with this manager, a date for deployment was arranged, with the plan
to have the robot in place in January 2024, during the busy time right after the
Christmas break when a number of new students arrive to start their degrees.

\subsubsection{Third Iteration}
\begin{figure}[t!]
    \centerline{\includegraphics[width=.9\columnwidth]{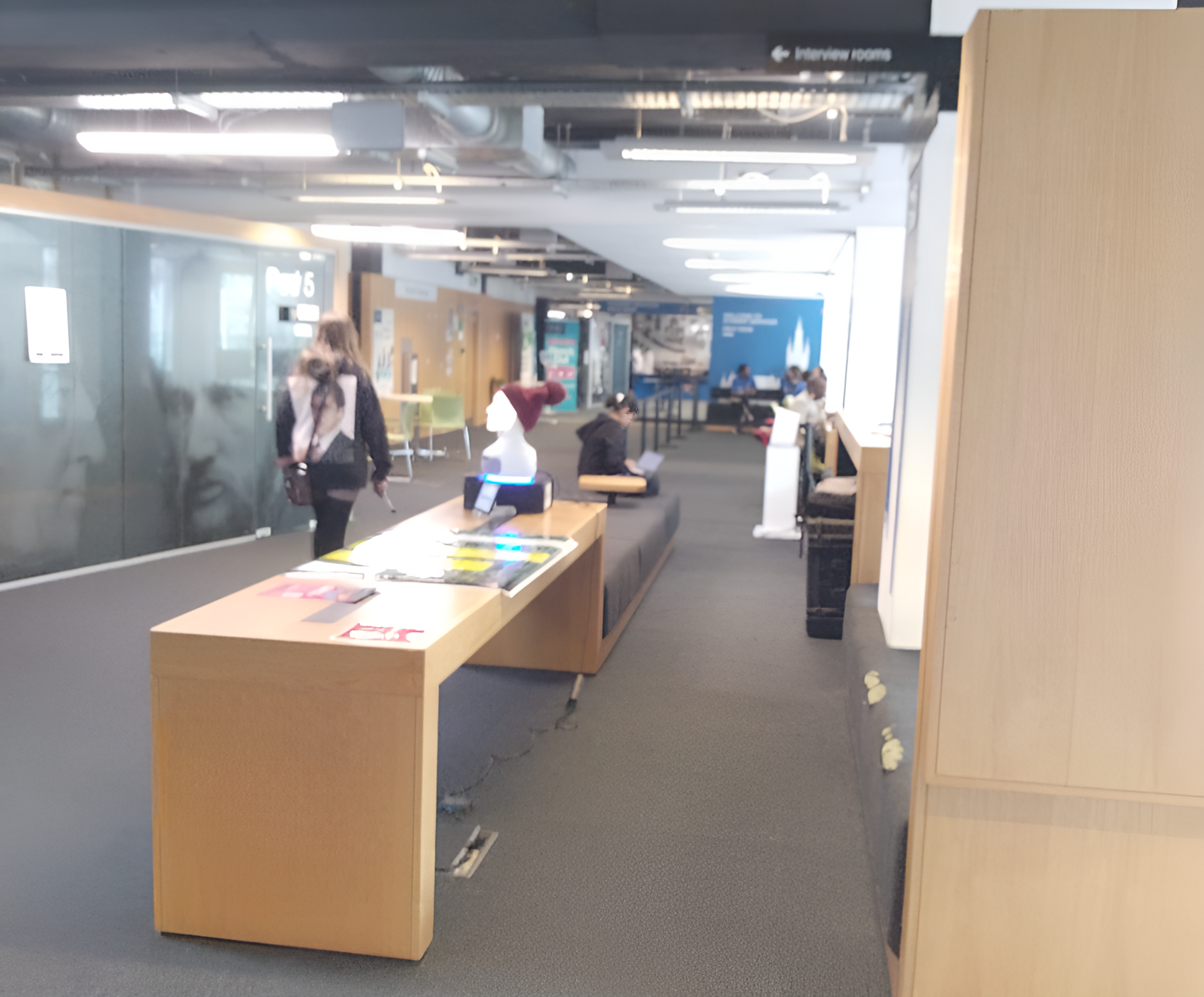}}
    \caption{The student support center}
    \label{fig:support-center}
\end{figure}

Shortly before the scheduled deployment was to begin, we came to know that there
had been another change of management, and agreements that we had with them and
the support staff were null and void. 
In an email, we were informed that they would no longer grant approval for the initiative, and the deployment was canceled.
Following this time, it took several months to get the deployment project back
on track and build a better relationship with the managers and the entire staff.
As part of this, we met online and presented videos of the Furhat robot, explaining that we did not require any resources except for a table to place the robot, and showed all possible dialogues that the robot might support. 
Following this consultation, we received full cooperation and support, which allowed us to carry out the field study just before the start of the 2024/25 academic year (Figure ~\ref{fig:support-center}).

\subsection{Refugees And Asylum Seekers Drop-In Service}

We collaborated with a Scottish charity supporting refugees, asylum seekers and migrant workers, where the first author has been volunteering with the drop-in service for the last 2 years. We decided to inquire about conducting a study at their space, as the relationship has already been established with the stakeholders due to the volunteer work.

Together with the community support staff we designed the study: aspects of discussion included the task itself, as well as the exact room where the robot would be deployed.

\subsubsection{The Public Space}
At first, we proposed a model similar to that used in the student support services, with the robot deployed at the time of the drop-in service alongside the staff in the same room. The staff was very concerned about that, mostly because they had not seen the robot in action and could not imagine how the drop-in service will work with the volunteers and the visitors. Their immediate answer was a negative one. As a response, we reached out to different staff members, asking to have a group meeting and presenting the robot to them, to which they agreed.

After the meeting, the staff felt more positive about the robot, but still preferred that we choose a different deployment space, such as their lounge, where people go to relax and can choose to interact with the robot in a less intense setting. We were happy with that, as the robot will still be deployed and used by the target population.

\subsubsection{The Task}

In that meeting, the staff observed the robot in action, giving information concerning local tax exemption, an example from the student support project. It helped them to come up with several ideas for the task. We wanted the robot to be genuinely useful, rather than simply presenting it with a predefined task based on our assumptions.

One of the managers suggested that the robot could help the refugees to enroll in English courses. However, this proposal turned out not to be feasible: the drop-in service staff were reluctant to give access to their database, and we were also unable to get support and cooperation from the colleges that offered the English language courses. 

The second iteration for the task concerned another gap the staff wanted us to bridge. A different charity, offering free hourly bike rentals, has collaborated with them, but the service users had significant difficulty in understanding the technical details of the process when presented in English. They wanted our multilingual robot to instruct and guide refugees and asylum seekers through technical tasks like downloading the app, locking or unlocking the bike, etc., and this was the final task that was agreed for the deployment which is currently ongoing.

%% file: sections/conclusions.tex
\section{Lessons learned}

Our experience of deploying social robots in public-facing environments has clearly
shown that technical readiness alone is not sufficient for success.

\begin{itemize}
    \item Stakeholders responded far more positively after seeing the robot in action; live
    demonstrations proved essential in overcoming initial skepticism.
    \item Staff turnover can significantly affect progress; clear documentation is crucial
    and plans for re-engagement should be prepared if necessary.
    \item Developing the use case for this sort of robot requires engaging with staff to 
    determine the needs of the users and the appropriate deployment location, 
    but should also be guided by technical and pragmatic considerations.
\end{itemize}